\documentclass[letterpaper]{article} 
\usepackage{aaai2026}  
\usepackage{times}  
\usepackage{helvet}  
\usepackage{courier}  
\usepackage[hyphens]{url}  
\usepackage{graphicx} 
\usepackage{amssymb}
\usepackage{natbib}  
\usepackage{caption} 
\usepackage{algorithm}
\usepackage{algorithmic}

\usepackage{newfloat}
\usepackage{listings}
\DeclareCaptionStyle{ruled}{labelfont=normalfont,labelsep=colon,strut=off} 
\floatstyle{ruled}
\newfloat{listing}{tb}{lst}{}
\floatname{listing}{Listing}

\usepackage{algorithm}
\usepackage{algorithmic}
\usepackage{multirow}
\usepackage{multicol}
\title{TextShield-R1: Reinforced Reasoning for Tampered Text Detection}
\author{
    Chenfan Qu\textsuperscript{\rm 1, 3}, Yiwu Zhong\textsuperscript{\rm 2}, Jian Liu\textsuperscript{\rm 3 *}, Xuekang Zhu\textsuperscript{\rm 3}, Bohan Yu\textsuperscript{\rm 3}, Lianwen Jin\textsuperscript{\rm 1 *}
}
\affiliations{
    \textsuperscript{\rm 1}South China University of Technology \\
    \textsuperscript{\rm 2} Peking University \\ \textsuperscript{\rm 3} Ant Group\\
    202221012612@mail.scut.edu.cn, rex.lj@antgroup.com, eelwjin@scut.edu.cn
}

\usepackage{bibentry}
\begin{document}

\maketitle

\begin{abstract}
The growing prevalence of tampered images poses serious security threats, highlighting the urgent need for reliable detection methods. Multimodal large language models (MLLMs) demonstrate strong potential in analyzing tampered images and generating interpretations.
However, they still struggle with identifying micro-level artifacts, exhibit low accuracy in localizing tampered text regions, and heavily rely on expensive annotations for forgery interpretation.
To this end, we introduce TextShield-R1, the first reinforcement learning based MLLM solution for tampered text detection and reasoning.
Specifically, our approach introduces Forensic Continual Pre-training, an easy-to-hard curriculum that well prepares the MLLM for tampered text detection by harnessing the large-scale cheap data from natural image forensic and OCR tasks.
During fine-tuning, we perform Group Relative Policy Optimization with novel reward functions to reduce annotation dependency and improve reasoning capabilities. 
At inference time, we enhance localization accuracy via OCR Rectification, a method that leverages the MLLM’s strong text recognition abilities to refine its predictions.
Furthermore, to support rigorous evaluation, we introduce the Text Forensics Reasoning (TFR) benchmark, comprising over 45k real and tampered images across 16 languages, 10 tampering techniques, and diverse domains. Rich reasoning-style annotations are included, allowing for comprehensive assessment. Our TFR benchmark simultaneously addresses seven major limitations of existing benchmarks and enables robust evaluation under cross-style, cross-method, and cross-language conditions.
Extensive experiments demonstrate that TextShield-R1 significantly advances the state of the art in interpretable tampered text detection.
\end{abstract}

\begin{links}
\link{Code}{https://github.com/qcf-568/TextShield}
\link{Datasets}{https://github.com/qcf-568/TextShield}
\link{Extended}{https://github.com/qcf-568/TextShield}
\end{links}

\begin{figure*}
\includegraphics[width=1.0\textwidth]{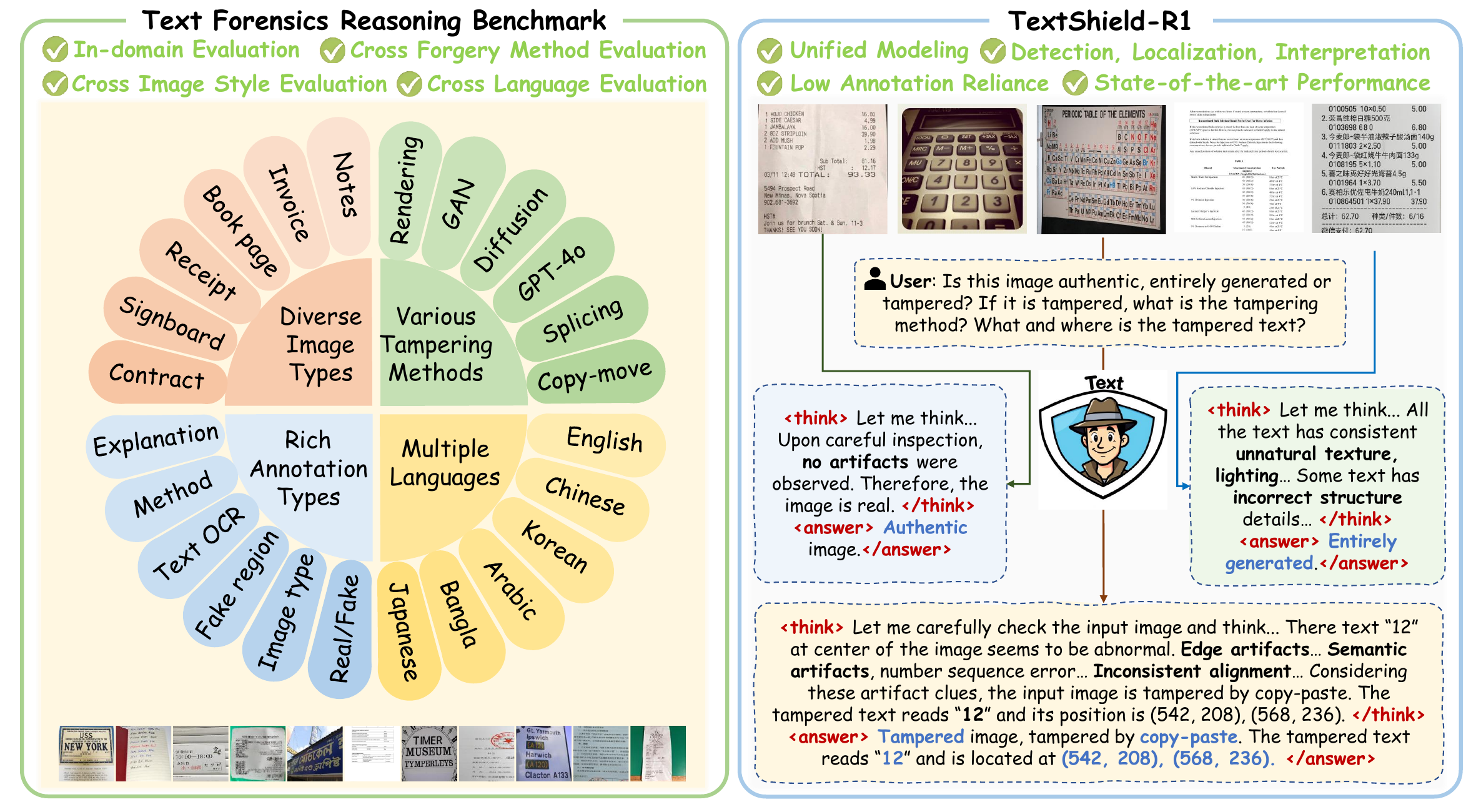} 
\setlength{\abovecaptionskip}{-0.3cm}
\captionof{figure}{We introduce the Text Forensics Reasoning benchmark, which features a wide range of image domains, diverse and up-to-date tampering methods, rich annotations, various languages and comprehensive out-of-distribution evaluation settings. We also propose TextShield-R1, the first reinforcement learning based model for tampered text detection.
}
\end{figure*}

\let\thefootnote\relax\footnotetext{* Corresponding Author}

\section{Introduction}
The rapid advancement of image processing technologies has greatly lowered the barrier to creating tampered text images. Unfortunately, such forgeries are increasingly exploited for fraud, rumor dissemination, and other malicious purposes, posing significant security threats~\cite{ffdn}. As a result, the reliable detection of tampered text has become a pressing research topic~\cite{ttd1}.

Recently, multimodal large language models (MLLMs) have demonstrated impressive human-like capabilities in perception and reasoning~\cite{segzero}, making them highly promising for multiple tasks~\cite{lan2024text4seg}. In the context of tampered text detection, MLLMs can analyze visual and semantic artifacts while also generating textual justifications for their predictions, thereby enhancing the interpretability and trustworthiness of their decisions. However, despite their potential, MLLMs still face several critical limitations that hinder their effectiveness in this domain.

First, \textbf{inadequate task alignment}. Existing base MLLMs are typically pretrained on macro perception tasks centered around high-level semantics, such as image captioning and object recognition. In contrast, tampered text detection demands a micro-level perception to discern semantic-agnostic artifacts. This significant discrepancy makes the task overly challenging for MLLMs, often leading to confusion and overfitting during fine-tuning on tampered texts.

Second, \textbf{heavy annotation dependence}. Most current MLLMs depend heavily on costly forgery interpretation annotations, which are typically obtained through expensive closed-source models such as GPT-4o. However, due to privacy concerns, many credential images (e.g., ID cards, contracts) with sensitive information are prohibited from external exposure. Moreover, since forgery artifacts are often unobvious, automatic annotation is error-prone and demands extensive manual cleaning. These challenges hinder large-scale training on real-world data. Even if full annotations are produced through labor-intensive efforts, the “spoon-fed teaching” nature of supervised fine-tuning can compromise the MLLM’s intrinsic reasoning and analytical capabilities.

Third, \textbf{poor localization accuracy}. MLLMs struggle to predict precise bounding boxes, especially for dense text. A naive solution involves integrating an extra traditional forgery localization model. However, this introduces additional latency. More importantly, such an approach can lead to misaligned predictions or excessive reliance on the localization model’s biased and unrobust outputs, thereby undermining the MLLM’s core aim of integration and universality.

To address these challenges and limitations, we propose \textbf{TextShield-R1} with innovations involving model pre-training, model fine-tuning, and model inference. 

\textbf{During pre-training}, we propose Forensic Continual Pre-training to address task misalignment. It is an easy-to-hard curriculum that begins by training the MLLM to detect tampered natural objects, which typically exhibit prominent and detectable artifacts. By leveraging large-scale, high-quality natural image forgery datasets, this approach also enables the model to acquire robust and generalizable forensic features. We further introduce 3D Forensic Learning, which enhances supervision across three complementary dimensions. While pre-training on tampered natural objects equips the MLLM with essential forgery detection capabilities, it inevitably compromises their OCR ability. To mitigate this trade-off, we interleave the curriculum with an OCR reference grounding task, resulting in a model that is both forensically aware and OCR capable for downstream tampered text detection task.

\textbf{During fine-tuning}, to reduce reliance on costly artifact interpretation annotations, we introduce the first reinforcement learning approach tailored for tampered text detection. Specifically, the model is trained using Group Relative Policy Optimization (GRPO), guided by carefully crafted reward functions. This reinforcement learning approach not only mitigates the need for extensive annotations but also enhances the MLLM’s reasoning capabilities.

\textbf{During inference}, to enhance the localization accuracy of tampered text, we introduce OCR Rectification. A task-specific OCR model first extracts both textual content and corresponding bounding box coordinates from the input image. Then, the MLLM predicts candidate tampered texts along with their associated bounding boxes. Each predicted box is refined by matching it to the OCR output based on content similarity and location proximity. If a suitable match is found, the OCR-derived box replaces the MLLM's original prediction. This approach effectively boosts localization performance by leveraging the MLLM’s strong text recognition capabilities.

In addition to our proposed method, we introduce the Text Forensics Reasoning (TFR) benchmark, a comprehensive high-quality resource for tampered text detection. Our TFR addresses all the seven critical limitations of existing benchmarks: a \textbf{limited domain} that focuses solely on document or scene text; a \textbf{narrow scope} that excludes entirely generated text images; an \textbf{unbalanced ratio} lacking real samples for false-positive evaluation; \textbf{insufficient diversity} of tampering techniques; \textbf{outdated tampering methods}; \textbf{insufficient out-of-distribution} evaluation settings; and \textbf{incomplete annotations} missing textual artifact interpretations.

Our TFR benchmark addresses all these shortcomings, enabling more rigorous and realistic evaluation of tampered text detection methods across domains, languages, and tampering styles.
Specifically, our benchmark comprises over 45k high-quality tampered text images and an equal number of real images with similar distributions. Notably, it is the first benchmark to comprehensively cover all three major text image types: documents, scene text, and ID-style cards. It is also the first to include both locally tampered and fully generated text images, capturing a broader spectrum of real-world forgery scenarios. It is the first to support robust evaluation across image styles, forgery methods, and languages simultaneously. Furthermore, our TFR benchmark stands out for its state-of-the-art forgery quality. For instance, it is the first to include realistic fake text images generated using GPT-4o.
To support deeper analysis, we also provide rich reasoning-style textual annotations, enabling more interpretable and LLM-compatible evaluations. We believe the TFR benchmark will serve as a valuable and foundational resource for advancing research in tampered text detection.

Extensive experiments conducted on the TFR benchmark and public benchmarks have validated our proposed method.

The main contributions of this paper are as follows:

\begin{itemize}
    \item \textbf{TextShield-R1}, the first reinforcement learning method for unified tampered text detection, effectively optimized through our carefully designed reward functions.
    \item \textbf{Forensic Continual Pre-training}, an easy-to-hard curriculum that well prepares MLLM for tampered text detection by harnessing the cheap data from other tasks.
    \item \textbf{OCR Rectification}, an innovative method that advances MLLM's localization performance with OCR results.
    \item \textbf{Text Forensics Reasoning benchmark}, a comprehensive high-quality benchmark that addresses all the seven critical issues of previous benchmarks.
\end{itemize}

\begin{table*}[]
\centering
\small
\setlength{\tabcolsep}{2.5pt}
\begin{tabular}{ccccccccccccccccccccccc}
\hline
\multirow{2}{*}{Dataset} &  & \multirow{2}{*}{\begin{tabular}[c]{@{}c@{}}Image\\ Domain\end{tabular}} &  & \multirow{2}{*}{\begin{tabular}[c]{@{}c@{}}Fake\\ Region\end{tabular}} &  & \multirow{2}{*}{\begin{tabular}[c]{@{}c@{}}Real\\ Num.\end{tabular}} &  & \multirow{2}{*}{\begin{tabular}[c]{@{}c@{}}Fake\\ Num.\end{tabular}} &  & \multirow{2}{*}{\begin{tabular}[c]{@{}c@{}}Method\\ Num.\end{tabular}} &  & \multirow{2}{*}{\begin{tabular}[c]{@{}c@{}}Lang\\ Num.\end{tabular}} &  & \multirow{2}{*}{\begin{tabular}[c]{@{}c@{}}Latest AIGC\\ Method (\#year)\end{tabular}} &  & \multicolumn{5}{c}{OOD Evaluation} &  & \multirow{2}{*}{\begin{tabular}[c]{@{}c@{}}Forgery\\ Explain.\end{tabular}} \\ \cline{17-21}
 &  &  &  &  &  &  &  &  &  &  &  &  &  &  &  & Style &  & Method &  & Lang. &  &  \\ \cline{1-1} \cline{3-3} \cline{5-5} \cline{7-7} \cline{9-9} \cline{11-11} \cline{13-13} \cline{15-15} \cline{17-17} \cline{19-19} \cline{21-21} \cline{23-23} 
T-SROIE &  & Doc. &  & Local &  & 0 &  & 920 &  & 1 &  & 1 &  & SR-Net (2019) &  &$\times$&  &$\times$&  &$\times$&  &$\times$\\
DocTamper &  & Doc. &  & Local &  & 0 &  & 170,000 &  & 3 &  & 2 &  & - &  &\checkmark&  &$\times$&  &$\times$&  &$\times$\\
RTM &  & Doc. &  & Local &  & 3000 &  & 6000 &  & 3 &  & 2 &  & - &  &$\times$&  &$\times$&  &$\times$&  &$\times$\\
Tampered-IC13 &  & S.T. &  & Local &  & 233 &  & 229 &  & 1 &  & 1 &  & SR-Net (2019) &  &$\times$&  &$\times$&  &$\times$&  &$\times$\\
OSTF &  & S.T. &  & Local &  & 4412 &  & 1980 &  & 8 &  & 1 &  & UDiffText (2023) &  &\checkmark&  &\checkmark&  &$\times$&  &$\times$\\
\multirow{2}{*}{Ours} &  & \multirow{2}{*}{\textbf{\begin{tabular}[c]{@{}c@{}}Doc.+S.T.\\ +Card\end{tabular}}} &  & \multirow{2}{*}{\textbf{\begin{tabular}[c]{@{}c@{}}Local+\\ Global\end{tabular}}} &  & \multirow{2}{*}{\textbf{45514}} &  & \multirow{2}{*}{45971} &  & \multirow{2}{*}{\textbf{10}} &  & \multirow{2}{*}{\textbf{16}} &  & \multirow{2}{*}{\textbf{GPT-4o (2025)}} &  & \multirow{2}{*}{\textbf{\checkmark}} &  & \multirow{2}{*}{\textbf{\checkmark}} &  & \multirow{2}{*}{\textbf{\checkmark}} &  & \multirow{2}{*}{\textbf{\checkmark}} \\
 & \multicolumn{1}{l}{} &  & \multicolumn{1}{l}{} &  & \multicolumn{1}{l}{} &  & \multicolumn{1}{l}{} &  & \multicolumn{1}{l}{} &  & \multicolumn{1}{l}{} &  & \multicolumn{1}{l}{} &  & \multicolumn{1}{l}{} &  & \multicolumn{1}{l}{} &  & \multicolumn{1}{l}{} &  & \multicolumn{1}{l}{} &  \\ \hline
\end{tabular}
\caption{Comparison between public text forensic benchmarks, which include T-SROIE~\cite{tsroie}, DocTamper~\cite{CVPR2023DocTamper}, RTM~\cite{luo2024RTM}, Tampered-IC13~\cite{wang2022tic13} and OSTF~\cite{ostf}. 'Doc.' denotes document, 'S.T.' denotes scene text, 'Num.' denotes number, 'Lang.' denotes language. 'OOD' includes out-of-distribution evaluation on unknown image styles, tampering methods and languages. 'Forgery Explain.' denotes textual forgery explanation annotation.}
\label{tab: datacomp}
\end{table*}

\begin{figure*}
    \includegraphics[width=1.0\textwidth]{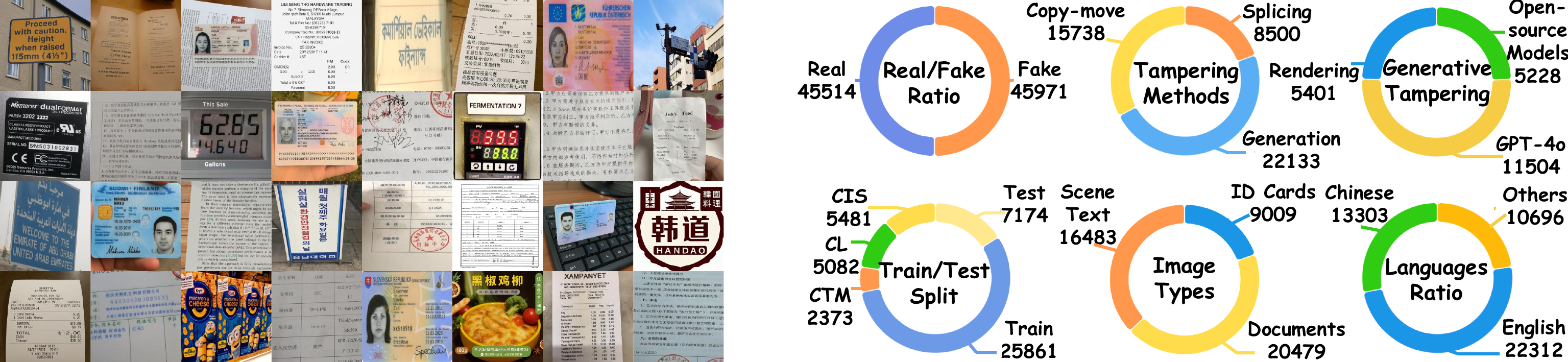}
    \caption{Representative samples (left) and data statistics (right) of the proposed Text Forensics Reasoning benchmark.}
    \label{fig: dataset}
\end{figure*}

\section{Related Works}

\subsection{Tampered Text Detection}
Early works in tampered text detection relied on handcrafted features or rules, such as printer classification~\cite{lampert2006printing} or template matching~\cite{ahmed2014forgery}. These methods are limited to specific layouts on scanned documents and do not work well on photographed text images~\cite{ttd3}. Some recent studies model tampered text detection as a semantic segmentation~\cite{dong2024robust, ttd2} or object detection~\cite{wang2022tic13, ostf} task. Despite progress, this prediction process remains a black box and produces unreliable results~\cite{qu2024textsleuth}. In addition, previous methods are tailored to a single image domain and cannot generalize across documents, scene texts, and ID cards. Some methods advanced the natural image forgery detection domain~\cite{li2021image, li2024unionformer, li2025toward, qu2024omni, zhang2025truemoe, zhu2025mesorch, su2025spase, qu2025webly, ma2025imdl, yu2024unified} but do not work well on text images~\cite{du2025forensichub, ditl2}.

\subsection{MLLM-Driven Image Forensics}
Recently, the rapid development of MLLM has accelerated progress in image forensics. FakeShield~\cite{xu2024fakeshield}, ForgeryGPT~\cite{li2024forgerygpt} and SIDA~\cite{huang2024sida} use MLLM to explain artifacts in natural images. M2F2-Det~\cite{M2F2-Det}, TTFG~\cite{ttfg} and AIGI-Holmes~\cite{zhou2025aigi} utilize MLLM to detect AI-generated face images. However, these methods depend heavily on forgery interpretation annotations. So-Fake~\cite{so-fake} leverages frozen SAM to detect AI-generated natural images. AvatarShield~\cite{avatarshield} harnesses temporal residual to detect AI-generated human videos. The above methods are designed for natural images. Due to the substantial differences between text forgeries and natural image forgeries~\cite{luo2024RTM}, existing methods are not effective for tampered text detection.

\section{Text Forensics Reasoning Benchmark}
\subsection{Motivation}
As shown in Table~\ref{tab: datacomp}, the construction of the Text Forensics Reasoning (TFR) is motivated by the seven crucial drawbacks of existing tampered text detection benchmarks:

\noindent $\bullet$ \textbf{Limited domain}: prior works are restricted to a single image domain (either document or scene text). 

\noindent $\bullet$ \textbf{Narrow scope}: all previous benchmarks fail to include entirely generated text images, which represent an increasingly common real-world scenario. 

\noindent $\bullet$ \textbf{Unbalanced ratio}: some benchmarks such as T-SROIE and DocTamper contain no real images, making it difficult to evaluate false positives effectively. 

\noindent $\bullet$ \textbf{Insufficient diversity}: only a small number of tampering techniques are considered (only one or three techniques in most works), reducing the robustness of model evaluations. 

\noindent $\bullet$ \textbf{Outdated quality}: Even the latest tampering method included in prior works is developed more than two years ago, lagging behind the rapid progress of forgery techniques.

\noindent $\bullet$ \textbf{Insufficient OOD evaluation}: existing benchmarks do not support thorough assessment of out-of-distribution (OOD) robustness, a crucial requirement for real-world deployment.

\noindent $\bullet$ \textbf{Incomplete annotations}: no public benchmark provides detailed textual reasoning annotations that analyze the visual and semantic artifacts of each tampered instance.

These crucial limitations significantly hinder the simulation of real-world scenarios and slow the development of practical text forensic methods.

\subsection{Construction}
We collected a diverse set of recent text images from document, scene text, and ID-style card domains. Image sources include the Internet and several public datasets. To create locally tampered text images, we applied copy-move, splicing, and rendering techniques using both manual efforts and meticulously designed automatic pipelines.
Additionally, we included local text forgeries generated by a GAN model (SR-Net) and advanced diffusion models (DiffUTE~\cite{chen2024diffute}, TextDiffuser-2~\cite{chen2024textdiffuser2} and UDiffText~\cite{zhao2024udifftext}). 
Entirely generated images are produced by querying GPT-4o-image-1, TextDiffuser-2~\cite{chen2024textdiffuser2}, AnyText-2~\cite{tuo2024anytext2}, and Control-Net~\cite{controlnet} with varied prompts. Manual filtering is conducted to ensure quality.

In total, the TFR benchmark comprises 45,971 fake images paired with 45,514 corresponding real images. Sample images are displayed on the left side of Figure~\ref{fig: dataset}. 
To enable thorough OOD robustness evaluation, we defined three additional sub-sets beyond the common in-domain test set:

\noindent $\bullet$ Cross-Image-Style (CIS): Text images from sources not present in the training set.

\noindent $\bullet$ Cross-Tampering-Method (CTM): Text images tampered using three methods (TextDiffuser-2, SR-Net, Control-Net) excluded from the training set.

\noindent $\bullet$ Cross-Language (CL): Text images in 10 languages that differ from those in the training set.

Besides providing high-quality forgeries, we also constructed reasoning-style forgery interpretation annotations for all images. These textual annotations were obtained by querying GPT-4o to analyze both visual and semantic artifacts, followed by manual refinement to ensure quality. Basic statistics are provided on the right side of Figure~\ref{fig: dataset}

\subsection{Highlights}

As shown in Table~\ref{tab: datacomp}, our TFR benchmark addresses all the seven major drawbacks of existing benchmarks and significantly outperforms them on multiple dimensions. 
For instance, our TFR features the most comprehensive coverage in terms of image domains (documents, scene text, and ID-style cards), forgery types (both local and global), tampering methods (10 techniques, including traditional methods and advanced AIGC methods), languages (16 types), out-of-distribution evaluation settings (3 distinct types), and annotation richness. This extensive scope establishes TFR as a valuable foundational resource for advancing text forensics.

\section{TextShield-R1}
\begin{figure}
    \includegraphics[width=0.48\textwidth]{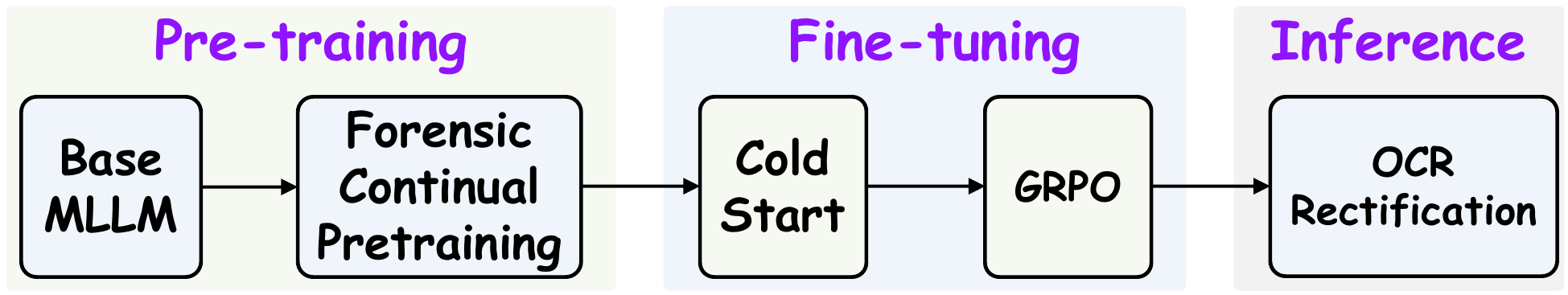}
    \caption{The overall pipeline of our TextShield-R1}
    \label{fig: pipeline}
\end{figure}

In this section we present TextShield-R1, a novel method for tampered text detection and reasoning, distinguished by advancements in both its training and inference pipelines.
As illustrated in Figure~\ref{fig: pipeline}, the training process begins with a \textbf{pre-training} stage involving Forensic Continual Pre-training on a base MLLM such as Qwen2.5-VL-7B. This pre-trained model then proceeds to a \textbf{fine-tuning} stage: initially with a small volume of fully annotated data to establish a cold start, followed by extensive fine-tuning with large-scale weakly annotated data using Group Relative Policy Optimization (GRPO). Here, "weakly annotated" implies that no artifact interpretation annotations are provided, requiring the model to reason about the artifact clues itself.
In the \textbf{inference} stage, we utilize OCR Rectification to enhance localization accuracy. TextShield-R1's plug-and-play design requires no architectural modifications to the base MLLM, ensuring its broad applicability across diverse MLLMs.

\subsection{Forensic Continual Pre-training}
Existing base MLLMs are primarily pretrained to recognize semantics at a macro level. In contrast, tampering detection necessitates identifying semantic-agnostic artifacts at a micro-level. Besides the pre-training gap, the tampered text detection task is considerably challenging in nature, yet high-quality training data remains costly and scarce. Consequently, directly fine-tuning an MLLM on this task inevitably leads to confusion and overfitting.

\begin{figure}
    \includegraphics[width=0.475\textwidth]{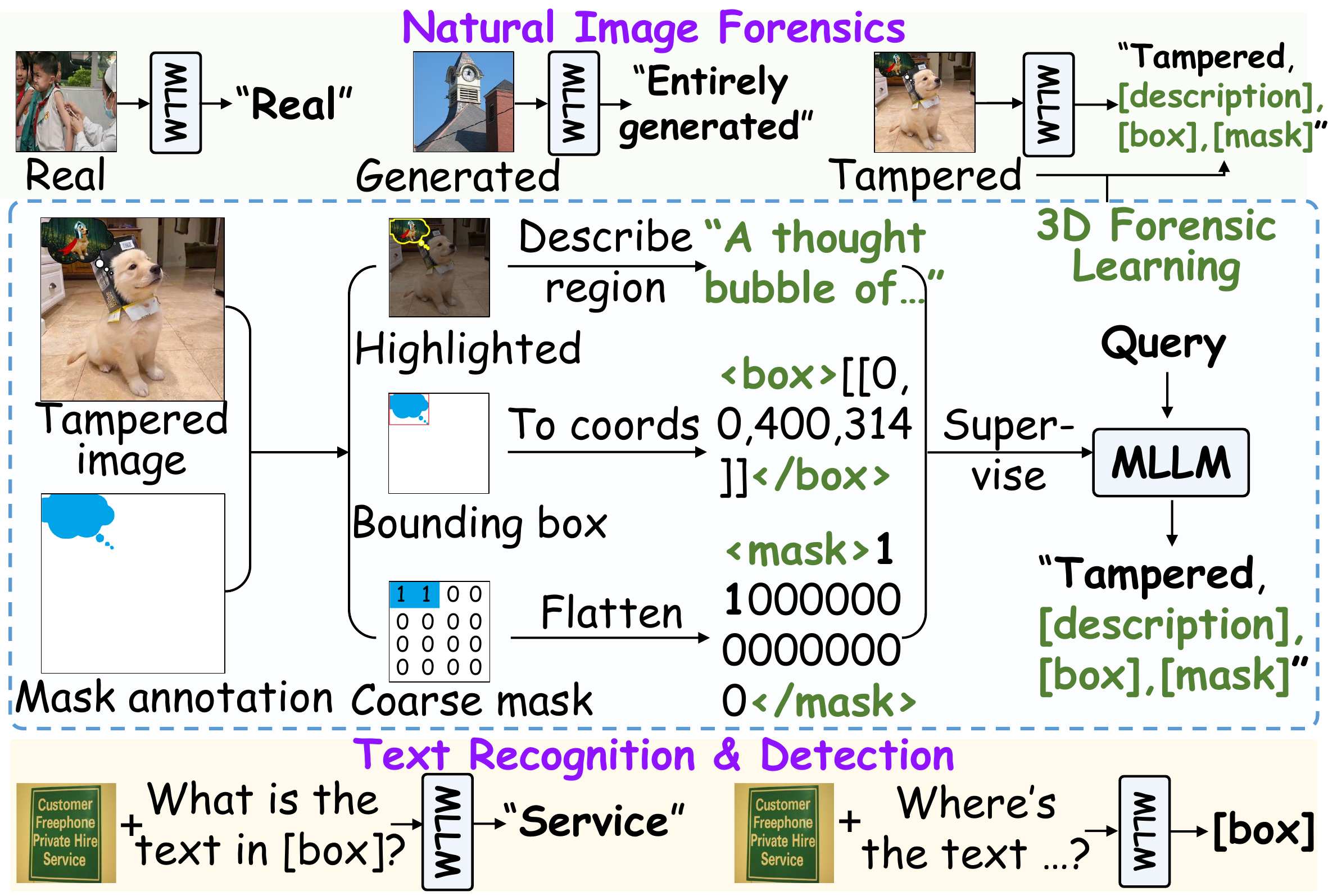}
    \caption{The Forensic Continual Pre-training pipeline. The MLLM is trained to distinguish between real, entirely generated, and locally tampered images. For locally tampered images, we introduce 3D Forensic Learning, which enhances supervision through three complementary dimensions. Additionally, we incorporate an OCR reference grounding task to prevent the forgetting of OCR-related knowledge.}
    \label{fig: pre-training}
\end{figure}

Inspired by the availability of large-scale, high-quality natural image forgeries, we propose to better prepare MLLMs for tampered text detection through continual pre-training on these datasets. Specifically, we pre-train the MLLM to classify whether an image is real, entirely generated, or locally tampered, as depicted in Figure~\ref{fig: pre-training}. 
For locally tampered natural images, we introduce a 3D Forensic Learning approach to enhance supervision through task collaboration. In addition to tampering classification, we require the MLLM to output the description, bounding box coordinates and mask string of the tampered object. 
The tampered object description annotation is generated by inputting the tampered images and masks into the Describe Anything Model~\cite{lian2025describe}. The bounding box coordinates are obtained by calculating the minimum bounding boxes of the tampered regions. The mask string is generated by interpolating a mask annotation to 32x32 pixels and representing each mask as a 0/1 string, where ’0’ denotes a real region and ’1’ denotes a tampered region.

Through pre-training on natural image forgeries, the MLLM's artifacts perception and grounding capabilities can be notably advanced. However, training on these non-text images inevitably erodes MLLM's previously acquired OCR knowledge, which is indispensable for tampered text detection. To overcome the dilemma, we interleave an OCR reference-grounding task: given a real text image and a randomly chosen text instance, the model is either (a) provided with the bounding box and asked to output the text, or (b) provided with the text and asked to output the bounding box. This dual task preserves the model’s OCR competence while reinforcing its forensic and grounding capabilities.

\subsection{Group Relative Policy Optimization}

Existing MLLMs depend heavily on textual forgery interpretation annotations, the creation of which is both costly and potentially raises privacy concerns. Furthermore, the traditional “spoon-fed teaching” supervised fine-tuning can dampen an MLLM’s native reasoning and analytical skills.

\begin{figure}
    \includegraphics[width=0.475\textwidth]{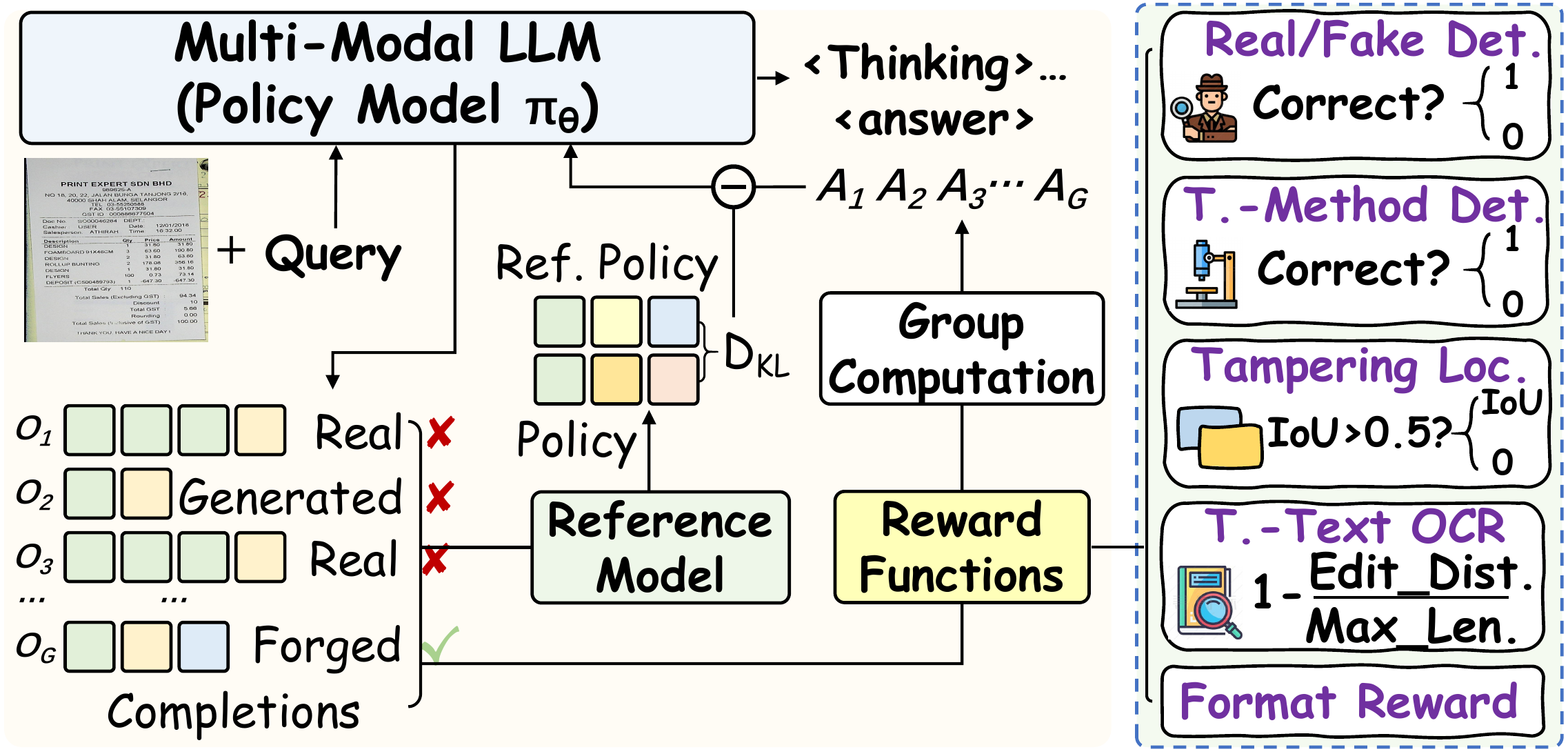}
    \caption{Under the GRPO framework, we optimize the model through five carefully designed reward functions.}
    \label{fig: grpo}
\end{figure}

Recent advances in reinforcement learning have demonstrated immense potential for guiding and optimizing LLMs, particularly following the advent of Group Relative Policy Optimization (GRPO). Under the GRPO framework, we design a set of task-specific reward functions to better instruct the model. As depicted in Figure~\ref{fig: grpo}, our method incorporates five distinct rewards: Real/Fake classification reward, forgery method detection reward, tampering localization reward, tampered text OCR reward and format reward.

\noindent \textbf{Real/Fake Classification Reward}: we encourage accurate image-level three-way classification of real, entirely generated and locally tampered images. A reward of 1 is assigned for the correct classification, and 0 otherwise.

\noindent \textbf{Forgery Method Detection Reward}: For tampered image, we encourage the model to identify whether a fake region was created by copy-paste or generation. A reward of 1 is assigned for the correct identification, and 0 otherwise. Different forgery methods often leave distinct artifacts, and this reward helps the model achieve more in-depth analysis for improved interpretation and generalization. 

\noindent \textbf{Tampering Localization Reward}: For tampered images, we encourage the model to accurately localize the tampered region. If the Intersection over Union (IoU) between the model’s prediction and the ground truth label exceeds 0.5, the reward score is set to the IoU value; otherwise, it is 0.

\noindent \textbf{Tampered Text OCR Reward}: For tampered images, we encourage the model to accurately recognize the tampered text. We utilize the normed Levenshtein distance to quantify the similarity between the model’s prediction and the ground truth string. The reward score is calculated as one minus the normed Levenshtein distance.

\noindent \textbf{Format Reward}: we encourage structured reasoning by rewarding outputs that embed reasoning within \(\langle think \rangle\)... \(\langle/ think \rangle\) and answers within \(\langle answer \rangle\)...\(\langle/ answer \rangle\) tags.

Through the proposed rewards, the model can effectively overcome its heavy reliance on textual annotations and fosters generalized analytical and reasoning skills.

\subsection{OCR Rectification}
While MLLMs excel at text recognition, they generally struggle with predicting precise text box coordinates. Given that the text detection task is considerably simpler for task-specific OCR models, and highly accurate OCR engines are readily available, we propose to leverage the MLLMs’ robust text recognition capabilities to refine their weak text localization predictions.

\begin{figure}
    \includegraphics[width=0.475\textwidth]{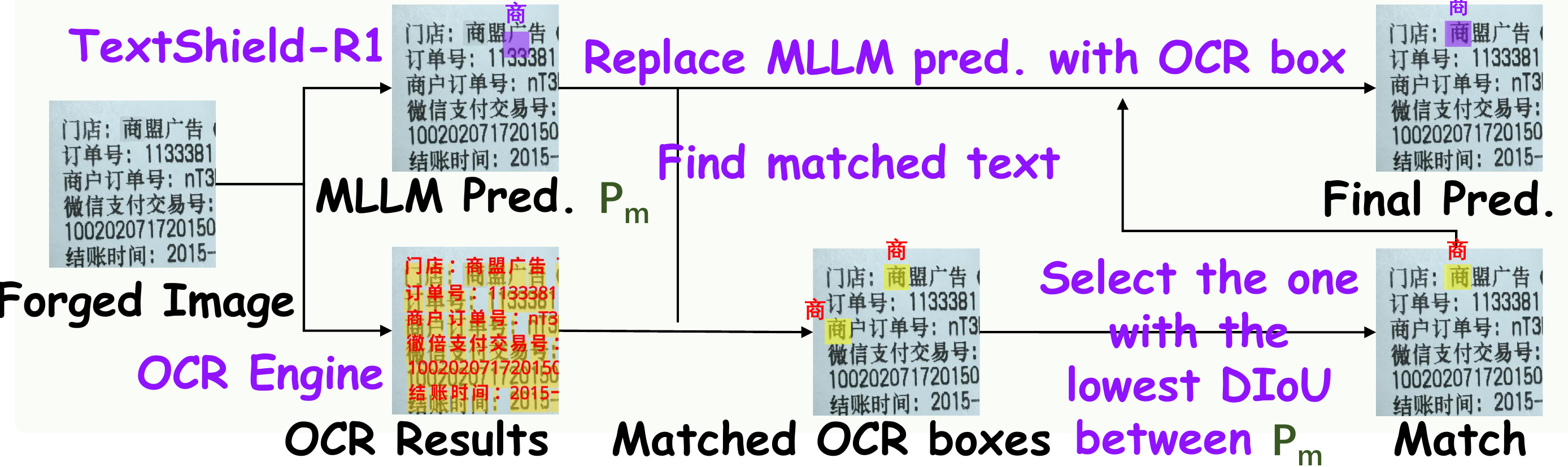}
    \caption{The proposed OCR Rectification pipeline,  illustrating the case when multiple matched texts exist.}
    \label{fig: ocr}
\end{figure}

The proposed approach is illustrated in Figure~\ref{fig: ocr}. Given an image predicted to contain tampering, we obtain OCR results from an OCR engine. These OCR results include textual content and bounding box coordinates for each detected text instance. Subsequently, for each predicted tampered text, we search the OCR results for a matched text instance. The matched text is defined as the instance exhibiting the minimum Levenshtein distance with the model’s prediction. If only one matched text is found, we directly replace the MLLM’s localization prediction with the bounding box provided by the OCR engine for that match. If multiple matched texts exist, we select the instance that maximizes the Distance IoU with the MLLM’s localization prediction. If no text instances satisfy the matching criterion (i.e., the normed Levenshtein distance between any OCR result and the predicted tampered text exceeds a fixed threshold of 0.2), we retain the MLLM’s original localization prediction.

Through our OCR Rectification, the issue of inaccurate text localization in MLLMs is significantly mitigated.

\section{Experiments}
\subsection{Implementation Details}
We adopt the Qwen2.5-VL-7B as the base MLLM of our TextShield-R1. In the Forensic Continual Pre-training stage, 120k locally tampered natural images are collected from CASIAv1v2~\cite{casia}, IMD20~\cite{imd20}, NIST16~\cite{nist16}, MIML~\cite{miml} datasets; 120k entirely generated natural forgeries are collected from the Community Forensic~\cite{communityforensic} dataset; 60k images from the COCO~\cite{coco} dataset and 60k images from the LAION~\cite{schuhmann2021laion} dataset are collected as the authentic images. The authentic text images from the training set of TFR benchmark are used for the OCR reference grounding task. We pre-train our model on the collected data for one epoch. The model is LoRA~\cite{hu2021lora} fine-tuned with LoRA rank 64 and is optimized by the AdamW optimizer with a learning rate decaying from 1e-4 to 0. In the fine-tuning stage, we train our model on the training set of the TFR benchmark. About 25\% fully annotated data are used at first to establish a cold start, the rest of the images are fine-tuned under GRPO.

\begin{table*}[ht!]
\centering
\small
\setlength{\tabcolsep}{2.5pt}
\begin{tabular}{ccccccccccccccccccccccccccccccccc}
\hline
\multirow{2}{*}{Method} &  & \multicolumn{7}{c}{Test set} &  & \multicolumn{7}{c}{CIS set} &  & \multicolumn{7}{c}{CTM set} &  & \multicolumn{7}{c}{CL set} \\ \cline{3-9} \cline{11-17} \cline{19-25} \cline{27-33} 
 &  & Cls. &  & OCR &  & Loc. &  & Res. &  & Cls. &  & OCR &  & Loc. &  & Res. &  & Cls. &  & OCR &  & Loc. &  & Res. &  & Cls. &  & OCR &  & Loc. &  & Res. \\ \hline
\multicolumn{33}{c}{Official pre-trained base MLLMs without fine-tuning} \\ \hline
GPT4o &  & 51.7 &  & 5.6 &  & 0.5 &  & 19.4 &  & 53.4 &  & 22.0 &  & 1.9 &  & 24.3 &  & 37.8 &  & 27.2 &  & 3.1 &  & 9.7 &  & 48.3 &  & 8.6 &  & 3.1 &  & 14.2 \\
MiniCPM\_V\_2.6 &  & 30.4 &  & 1.6 &  & 0.0 &  & 3.2 &  & 31.4 &  & 4.8 &  & 0.0 &  & 3.2 &  & 26.2 &  & 7.0 &  & 0.0 &  & 2.5 &  & 30.9 &  & 0.4 &  & 0.0 &  & 1.9 \\
InternVL3-2B &  & 33.2 &  & 5.4 &  & 0.0 &  & 8.5 &  & 40.0 &  & 14.0 &  & 0.1 &  & 9.9 &  & 20.3 &  & 18.2 &  & 0.4 &  & 4.0 &  & 34.5 &  & 6.7 &  & 0.0 &  & 6.9 \\
InternVL3-8B &  & 40.4 &  & 9.3 &  & 0.2 &  & 17.9 &  & 47.0 &  & 20.9 &  & 0.8 &  & 20.3 &  & 25.2 &  & 31.5 &  & 1.7 &  & 8.8 &  & 45.1 &  & 10.3 &  & 0.5 &  & 18.1 \\
Qwen2.5-VL-3B &  & 46.2 &  & 1.8 &  & 0.1 &  & 9.5 &  & 48.4 &  & 5.9 &  & 0.3 &  & 15.3 &  & 42.2 &  & 7.7 &  & 0.2 &  & 5.6 &  & 47.1 &  & 1.5 &  & 0.2 &  & 7.9 \\
Qwen2.5-VL-7B &  & 42.6 & \multicolumn{1}{l}{} & 6.4 & \multicolumn{1}{l}{} & 0.1 & \multicolumn{1}{l}{} & 9.5 &  & 49.9 & \multicolumn{1}{l}{} & 19.4 & \multicolumn{1}{l}{} & 0.4 & \multicolumn{1}{l}{} & 17.6 &  & 34.0 & \multicolumn{1}{l}{} & 21.0 & \multicolumn{1}{l}{} & 0.6 & \multicolumn{1}{l}{} & 4.5 &  & 50.1 & \multicolumn{1}{l}{} & 11.1 & \multicolumn{1}{l}{} & 0.2 & \multicolumn{1}{l}{} & 10.4 \\ \hline
\multicolumn{33}{c}{MLLMs fine-tuned with full training set images} \\ \hline
MiniCPM\_V\_2.6 &  & 76.2 &  & 17.6 &  & 11.2 &  & 41.1 &  & 71.7 &  & 24.5 &  & 22.8 &  & 32.3 &  & 64.8 &  & 20.7 &  & 24.9 &  & 27.3 &  & 81.5 &  & 33.6 &  & 20.5 &  & 40.3 \\
InternVL3-2B &  & 75.4 &  & 18.1 &  & 10.3 &  & 40.6 &  & 68.5 &  & 23.1 &  & 21.4 &  & 32.0 &  & 62.5 &  & 18.7 &  & 26.2 &  & 25.0 &  & 80.2 &  & 31.7 &  & 21.0 &  & 39.5 \\
InternVL3-8B &  & 78.6 &  & 21.9 &  & 15.4 &  & 41.7 &  & 70.8 &  & 27.6 &  & 25.2 &  & 33.8 &  & 67.7 &  & 23.6 &  & 31.4 &  & 32.3 &  & 84.3 &  & 38.0 &  & 24.8 &  & 42.0 \\
Qwen2.5-VL-3B &  & 77.5 &  & 18.6 &  & 11.6 &  & 42.9 &  & 72.3 &  & 25.0 &  & 20.9 &  & 33.6 &  & 63.0 &  & 18.8 &  & 25.6 &  & 24.6 &  & 80.9 &  & 32.4 &  & 21.4 &  & 39.7 \\
Qwen2.5-VL-7B &  & 79.1 &  & 24.3 &  & 18.2 &  & 42.9 &  & 71.1 &  & 30.7 &  & 26.5 &  & 35.7 &  & 73.6 &  & 26.3 &  & 34.2 &  & 36.2 &  & 85.1 &  & 38.2 &  & 25.5 &  & 43.1 \\
FakeShield &  & 70.5 &  & 9.2 &  & 5.4 &  & 35.6 &  & 62.0 &  & 14.8 &  & 10.6 &  & 29.2 &  & 57.5 &  & 15.1 &  & 17.3 &  & 21.4 &  & 70.3 &  & 23.9 &  & 11.3 &  & 34.8 \\
FakeShield* &  & 79.1 &  & 24.3 &  & 7.6 &  & 42.8 &  & 71.1 &  & 30.5 &  & 15.0 &  & 35.6 &  & 73.6 &  & 26.3 &  & 21.8 &  & 36.2 &  & 85.1 &  & 38.1 &  & 15.6 &  & 42.9 \\
SIDA &  & 71.2 &  & 9.2 &  & 5.6 &  & 35.7 &  & 62.2 &  & 14.9 &  & 10.8 &  & 29.4 &  & 57.5 &  & 15.1 &  & 17.3 &  & 21.5 &  & 70.4 &  & 23.8 &  & 11.5 &  & 25.0 \\
SIDA* &  & 79.2 &  & 24.3 &  & 7.7 &  & 42.9 &  & 71.4 &  & 30.9 &  & 15.1 &  & 35.7 &  & 73.6 &  & 26.3 &  & 21.8 &  & 36.3 &  & 85.2 &  & 38.2 &  & 15.8 &  & 43.0 \\
Ours &  & \textbf{88.1} & \textbf{} & \textbf{47.6} & \textbf{} & \textbf{57.8} &  & \textbf{58.8} &  & \textbf{72.9} & \textbf{} & \textbf{62.1} & \textbf{} & \textbf{61.0} & \textbf{} & \textbf{56.5} &  & \textbf{88.8} & \textbf{} & \textbf{45.6} & \textbf{} & \textbf{68.3} & \textbf{} & \textbf{51.2} &  & \textbf{85.5} & \textbf{} & \textbf{39.0} &  & \textbf{40.6} & \textbf{} & \textbf{46.2} \\ \hline
\end{tabular}
\caption{Comparison experiments. 'Cls' denotes the real/generated/tampered task with the accuracy metric. 'OCR' denotes the tampered text recognition task with the OCR accuracy metric. 'Loc.' denotes the tampered text localization task with the IoU metric. 'Res.' denotes the forgery reasoning task, using the average score of cosine similarity, Rouge-L and BLEU as metric. 'FakeShield*', 'SIDA*' denote FakeShield~\cite{xu2024fakeshield} and SIDA~\cite{huang2024sida} with the Qwen2.5-VL-7B as MLLM.}
\label{tab: comp}
\end{table*}

\begin{table*}[ht!]
\centering
\small
\setlength{\tabcolsep}{2.4pt}
\begin{tabular}{ccccccccccccccccccccccccccccccccccc}
\hline
\multirow{2}{*}{Num} &  & \multirow{2}{*}{Ablation} &  & \multicolumn{7}{c}{Test set} &  & \multicolumn{7}{c}{CIS set} &  & \multicolumn{7}{c}{CTM set} &  & \multicolumn{7}{c}{CL set} \\ \cline{5-11} \cline{13-19} \cline{21-27} \cline{29-35} 
 &  &  &  & Cls. &  & OCR &  & Loc. &  & Res. &  & Cls. &  & OCR &  & Loc. &  & Res. &  & Cls. &  & OCR &  & Loc. &  & Res. &  & Cls. &  & OCR &  & Loc. &  & Res. \\ \cline{1-5} \cline{7-7} \cline{9-9} \cline{11-11} \cline{13-13} \cline{15-15} \cline{17-17} \cline{19-19} \cline{21-21} \cline{23-23} \cline{25-25} \cline{27-27} \cline{29-29} \cline{31-31} \cline{33-33} \cline{35-35} 
(1) &  & Baseline &  & 79.1 &  & 24.3 &  & 18.2 &  & 42.9 &  & 71.1 &  & 30.7 &  & 26.5 &  & 35.7 &  & 73.6 &  & 26.3 &  & 34.2 &  & 36.2 &  & 85.1 &  & 38.2 &  & 25.5 &  & 43.1 \\
(2) &  & w.o. FCP &  & 75.8 &  & 21.9 &  & 12.7 &  & 39.0 &  & 68.4 &  & 25.0 &  & 20.9 &  & 30.6 &  & 66.3 &  & 23.7 &  & 25.5 &  & 30.1 &  & 83.9 &  & 38.5 &  & 26.0 &  & 41.8 \\
(3) &  & w.o. GRPO &  & 87.6 &  & 46.8 &  & 57.7 &  & 58.6 &  & 72.3 &  & 61.7 &  & 60.8 &  & 56.2 &  & 88.1 &  & 45.3 &  & 68.2 &  & 50.9 &  & 85.4 &  & 38.5 &  & 40.2 &  & 46.1 \\
(4) &  & w.o. OCR Rect. &  & \textbf{88.1} & \textbf{} & \textbf{47.6} & \textbf{} & 42.7 &  & \textbf{58.8} &  & \textbf{72.9} & \textbf{} & \textbf{62.1} & \textbf{} & 56.6 & \textbf{} & \textbf{56.5} &  & \textbf{88.8} & \textbf{} & \textbf{45.6} & \textbf{} & 57.9 & \textbf{} & \textbf{51.2} &  & \textbf{85.5} & \textbf{} & \textbf{39.0} &  & 32.3 & \textbf{} & \textbf{46.2} \\
(5) &  & TextShield-R1 &  & \textbf{88.1} & \textbf{} & \textbf{47.6} & \textbf{} & \textbf{57.8} &  & \textbf{58.8} &  & \textbf{72.9} & \textbf{} & \textbf{62.1} & \textbf{} & \textbf{61.0} & \textbf{} & \textbf{56.5} &  & \textbf{88.8} & \textbf{} & \textbf{45.6} & \textbf{} & \textbf{68.3} & \textbf{} & \textbf{51.2} &  & \textbf{85.5} & \textbf{} & \textbf{39.0} &  & \textbf{40.6} & \textbf{} & \textbf{46.2} \\ \hline
\end{tabular}
\caption{Ablation study on the proposed modules. 'w.o.' denotes 'without'. 'FCP' denotes the Forensic Continual Pre-training approach. 'OCR Rect.' denotes the proposed OCR Rectification.}
\label{tab: ablmain}
\end{table*}

\begin{table*}[ht!]
\centering
\small
\setlength{\tabcolsep}{2pt}
\begin{tabular}{ccccccccccccccccccccccccccccccccccccccc}
\hline
\multirow{2}{*}{Num.} &  & \multicolumn{5}{c}{Ablations} &  & \multicolumn{7}{c}{Test set} &  & \multicolumn{7}{c}{CIS set} &  & \multicolumn{7}{c}{CTM set} &  & \multicolumn{7}{c}{CL set} \\ \cline{3-7} \cline{9-15} \cline{17-23} \cline{25-31} \cline{33-39} 
 &  & Nat. &  & 3D-FL &  & OCR &  & Cls. &  & OCR &  & Loc. &  & Res. &  & Cls. &  & OCR &  & Loc. &  & Res. &  & Cls. &  & OCR &  & Loc. &  & Res. &  & Cls. &  & OCR &  & Loc. &  & Res. \\ \cline{1-1} \cline{3-3} \cline{5-5} \cline{7-7} \cline{9-9} \cline{11-11} \cline{13-13} \cline{15-15} \cline{17-17} \cline{19-19} \cline{21-21} \cline{23-23} \cline{25-25} \cline{27-27} \cline{29-29} \cline{31-31} \cline{33-33} \cline{35-35} \cline{37-37} \cline{39-39} 
(1) &  & $\times$ &  & $\times$ &  & $\times$ &  & 75.8 &  & 21.9 &  & 12.7 &  & 39.0 &  & 68.4 &  & 25.0 &  & 20.9 &  & 30.6 &  & 66.3 &  & 23.7 &  & 25.5 &  & 30.1 &  & 83.9 &  & 38.5 &  & 26.0 &  & 41.8 \\
(2) &  & \checkmark &  & $\times$ &  & $\times$ &  & 80.9 &  & 12.7 &  & 9.8 &  & 34.0 &  & 65.1 &  & 14.3 &  & 12.4 &  & 27.5 &  & 57.6 &  & 16.9 &  & 15.1 &  & 26.2 &  & 80.1 &  & 18.6 &  & 17.4 &  & 36.5 \\
(3) &  & \checkmark &  & \checkmark &  & $\times$ &  & 82.3 &  & 11.5 &  & 13.9 &  & 39.2 &  & 67.7 &  & 12.0 &  & 15.8 &  & 29.8 &  & 63.6 &  & 14.7 &  & 19.1 &  & 28.4 &  & 81.5 &  & 9.9 &  & 22.9 &  & 37.2 \\
(4) &  & \checkmark &  & $\times$ &  & \checkmark &  & 83.2 &  & 40.9 &  & 48.6 &  & 52.7 &  & 70.4 &  & 56.8 &  & 52.0 &  & 51.4 &  & 78.6 &  & 41.1 &  & 56.1 &  & 48.3 &  & 82.5 &  & 37.8 &  & 34.2 &  & 43.0 \\
(5) &  & \checkmark &  & \checkmark &  & \checkmark &  & \textbf{88.1} & \textbf{} & \textbf{47.6} & \textbf{} & \textbf{57.8} &  & \textbf{58.8} &  & \textbf{72.9} & \textbf{} & \textbf{62.1} & \textbf{} & \textbf{61.0} & \textbf{} & \textbf{56.5} &  & \textbf{88.8} & \textbf{} & \textbf{45.6} & \textbf{} & \textbf{68.3} & \textbf{} & \textbf{51.2} &  & \textbf{85.5} & \textbf{} & \textbf{39.0} &  & \textbf{40.6} & \textbf{} & \textbf{46.2} \\ \hline
\end{tabular}
\caption{Ablation study on the proposed Forensic Continual Pre-training method. 'Nat.' denotes pre-training the model to identify whether a natural image is real/generated/tampered. '3D-FL' denotes the proposed 3D Forensic Learning approach. 'OCR' denotes including the OCR reference grounding task.}
\label{tab: ablfcd}
\end{table*}

\subsection{Comparison Study}
We compare our method with both the official pretrained versions and the TFR fine-tuned versions of MiniCPMV2.6~\cite{yao2024minicpm}, Qwen2.5-VL-3B~\cite{bai2025qwen2.5vl}, Qwen2.5-VL-7B, InternVL3-2B~\cite{zhu2025internvl3} and InternVL3-8B. The results are shown in Table~\ref{tab: comp}. Evidently, all the pre-trained models have low scores. This validates that the tampered text detection task is challenging and cannot be well solved by existing MLLMs. Additionally, our TextShield-R1 establishes a new state-of-the-art in the field of MLLM-based tampered text detection, which confirms the effectiveness of the proposed method in all the image-level classification, tampered text recognition, tampered text localization and artifacts reasoning tasks.

\subsection{Ablation Study}
Table~\ref{tab: ablmain} presents an ablation study of our proposed modules. Setting (1) is the Qwen2.5-VL-7B baseline, which includes none of our proposed modules, while setting (5) is our full TextShield-R1 model incorporating all of them. Removing Forensic Continual Pre-training (setting (2)) from the full model results in a significant performance degradation, with scores falling even below the baseline (1). This highlights the critical role of this pre-training stage, which establishes a foundational understanding of the complex text image forensics task. Without it, the model struggles to learn effectively and converge during the subsequent GRPO fine-tuning.
The full model (5) performs comparably to setting (3) despite using textual forgery reasoning annotations for only a quarter of the training data. This result validates that our novel reward functions effectively enable the model to learn forgery reasoning from partially annotated datasets. Furthermore, by integrating our OCR Rectification method, the full model (5) outperforms setting (4) on the tampered text localization task across all four test sets. This confirms that OCR Rectification enhances localization performance by effectively leveraging the inherent OCR strengths of the MLLM.

Table~\ref{tab: ablfcd} details the ablation study for our Forensic Continual Pre-training stage. Setting (1) is the baseline model without any continual pre-training. Setting (2), which involves pre-training solely on distinguishing between real, generated, and tampered images, improves image-level classification but leads to a significant drop in OCR and localization performance. This occurs because this narrow pre-training objective causes catastrophic forgetting of the model's inherent OCR and localization capabilities. Adding the 3D Forensic Learning task (setting (3)) notably improves localization performance, though the OCR score remains low due to the continued forgetting of text recognition knowledge. Our final pre-trained model (setting (5)) is achieved by adding the OCR Reference Grounding task, which makes the model both forensically aware and OCR-capable. Finally, the superior performance of setting (5) over setting (4) demonstrates that 3D Forensic Learning is essential, as it helps the model learn generalized features for forgery localization.

\section{Conclusion}
In this work, we introduced TextShield-R1, a novel framework that systematically addresses key challenges in MLLM-based tampered text detection. Our Forensic Continual Pre-training bridges the gap between general-purpose pre-training and fine-grained forensic analysis using an easy-to-hard curriculum. To reduce reliance on expensive annotations and foster deeper analytical skills, we pioneered a reinforcement learning approach, which guides the model with novel reward functions. Furthermore, our OCR Rectification method elegantly resolves poor localization accuracy by leveraging the MLLM’s own powerful text recognition capabilities to refine its predictions. We also constructed the Text Forensics Reasoning (TFR) benchmark. This comprehensive resource remedies seven major deficiencies of prior datasets by incorporating diverse image domains, modern forgery techniques, and robust cross-domain, cross-method, and cross-language test settings. Extensive experiments validate that TextShield-R1 significantly advances the state of the art in detection accuracy, generalization, and interpretability. By tackling critical gaps in both methodology and evaluation, our work provides a robust foundation for future research into developing more reliable and trustworthy forensic AI systems.

\section{Acknowledgments}
This research is supported in part by the National Natural Science Foundation of China (Grant No.:62476093) and Ant Group Research Intern Program.

\bibliography{aaai2026}

\end{document}